\documentclass[runningheads]{llncs}

\usepackage[T1]{fontenc}
\def\doi#1{\href{https://doi.org/\detokenize{#1}}{\url{https://doi.org/\detokenize{#1}}}}
\usepackage{graphicx}
\usepackage{listings}
\usepackage[english]{babel}
\usepackage{amsmath}
\usepackage[colorlinks=true, allcolors=black]{hyperref}

\usepackage{booktabs} % tables
\usepackage{multirow} % tables
\usepackage[table]{xcolor} % table cell and font color
\usepackage{array}
\usepackage{bbm}      % indicator function 

\begin{document}
%
%\title{Contribution Title\thanks{Supported by organization x.}}
\title{AttentionHTR: Handwritten Text Recognition Based on Attention Encoder-Decoder Networks}
\titlerunning{AttentionHTR}
\author{Dmitrijs Kass\inst{1} \and
Ekta Vats\inst{2}\orcidID{0000-0003-4480-3158}}
\authorrunning{D. Kass and E. Vats}
\institute{Department of Information Technology, Uppsala University, Sweden
\email{dmitrijs.kass@it.uu.se}\\
 \and
Centre for Digital Humanities, Department of ALM, Uppsala University, Sweden\\
\email{ekta.vats@abm.uu.se}}
\maketitle              % typeset the header of the contribution
\begin{abstract}
This work proposes an attention-based sequence-to-sequence model for handwritten word recognition and explores transfer learning for data-efficient training of HTR systems. To overcome training data scarcity, this work leverages models pre-trained on scene text images as a starting point towards tailoring the handwriting recognition models. ResNet feature extraction and bidirectional LSTM-based sequence modeling stages together form an encoder. The prediction stage consists of a decoder and a content-based attention mechanism. The effectiveness of the proposed end-to-end HTR system has been empirically evaluated on a novel multi-writer dataset Imgur5K and the IAM dataset. The experimental results evaluate the performance of the HTR framework, further supported by an in-depth analysis of the error cases. Source code and pre-trained models are available at GitHub\footnote{\href{https://github.com/dmitrijsk/AttentionHTR}{https://github.com/dmitrijsk/AttentionHTR}}.
%A Scene Text Recognition (STR) benchmark model trained on synthetic scene text images is therefore used to perform transfer learning from STR domain to HTR.

%Handwritten text possesses high variability due to different writing styles, languages, and scripts. In order to have an accurate and robust Handwritten Text Recognition (HTR) system, it is crucial to perform training on a variety of data, in a large amount. However, due to the limited availability of annotated training data, especially for multi-writer texts, the performance of HTR systems is compromised. 

\end{abstract}
\section{Introduction}
Historical archives and cultural institutions contain rich heritage collections from historical times that are to be digitized to prevent degradation over time. The importance of digitization has led to a strong research interest in designing methods for automatic handwritten text recognition (HTR). However, handwritten text possesses variability in handwriting styles, and the documents are often heavily degraded. Such issues render the digitization of handwritten material more challenging and suggest the need to have more sophisticated HTR systems.

The current state-of-the-art in HTR is dominated by deep learning-based methods that require a significant amount of training data. Further, to accurately model the variability in writing styles in a multi-writer scenario, it is important to train the neural network on a variety of handwritten texts. However, only a limited amount of annotated data is available to train an end-to-end HTR model from scratch, and that affects the performance of the HTR system. This work proposes an end-to-end HTR system based on attention encoder-decoder architecture and presents a transfer learning-based approach for data-efficient training. In an effort toward designing a data-driven HTR pipeline, a Scene Text Recognition (STR) benchmark model \cite{baek2019wrong} is studied, which is trained on nearly 14 million synthetic scene text word images. The idea is to fine-tune the STR benchmark model on different handwritten word datasets.

%The contributions of this work are as follows. First, a novel end-to-end HTR system is presented, trained on a wide variety of texts. Second, to overcome the issue of training data scarcity, this work leverages pre-trained models trained on synthetic scene text images and explores transfer learning from the STR domain to HTR. Experimental results highlight the importance of using transfer learning and suggest significant improvement in word recognition accuracy. Third, the focus of this work is to recognize words written by a variety of writers, and therefore Imgur5K dataset \cite{krishnan2021textstylebrush} is used with data from 5000 writers, and also the benchmark multi-writer IAM handwriting database \cite{marti2002iam}. Furthermore, this work presents a comparison of different approaches to fine-tuning. Lastly, the paper presents a comprehensive error analysis using different techniques and suggests scope for improvements.

The novelty and technical contributions of this work are as follows. (a) The goal of this work is to leverage transfer learning to enable HTR in cases where training data is scarce. (b) Scene text images and a new dataset, Imgur5K \cite{krishnan2021textstylebrush}, are used for transfer learning. The Imgur5K dataset contains a handwritten text by approximately 5000 writers, which allows our proposed model to generalize better on unseen examples. (c) Our transfer learning-based framework produces a model that is applicable in the real world as it was trained on word examples from thousands of authors, with varying imaging conditions. (d) The proposed attention-based architecture is simple, modular, and reproducible, more data can be easily added in the pipeline, further strengthening the model’s accuracy. (e) A comprehensive error analysis using different techniques such as bias-variance analysis, character-level analysis, and visual analysis is presented. (f) An ablation study is conducted to demonstrate the importance of the proposed framework.

%Seq2Seq is a type of Encoder-Decoder model using RNN. 

\section{Related work}
\label{sec:literature}

In document analysis literature, popular approaches towards handwriting recognition include Hidden Markov Models \cite{rodriguez2012model}, Recurrent Neural Networks (RNN) \cite{frinken2012novel}, CNN-RNN hybrid architectures \cite{dutta2018improving}, and attention-based sequence to sequence (seq2seq) models \cite{bluche2017scan,kang2018convolve,sueiras2018offline}. Recent developments \cite{kang2021candidate,michael2019evaluating} suggest that the current state-of-the-art in HTR is advancing towards using attention-based seq2seq models. RNNs are commonly used to model the temporal nature of the text, and the attention mechanism performs well when used with RNNs to focus on the useful features at each time step \cite{kang2018convolve}.

Seq2seq models follow an encoder-decoder architecture, where one network encodes an input sequence into a fixed-length vector representation, and the other decodes it into a target sequence. However, using a fixed-length vector affects the performance, and therefore attention mechanisms are gaining importance due to enabling an automatic search for the most relevant elements of an input sequence to predict a target sequence \cite{bahdanau2014neural}. 

This work investigates attention-based encoder-decoder networks \cite{baek2019wrong} for handwriting recognition, based on ResNet for feature extraction, bidirectional LSTMs for sequence modeling, a content-based attention mechanism for prediction \cite{bahdanau2014neural}, and transfer learning from STR domain to HTR to overcome lack of training data. To this end, the STR benchmark \cite{baek2019wrong} has been studied, and the pre-trained models from STR are used in this work to fine-tune the HTR models. In \cite{baek2019wrong}, the models were trained end-to-end on the union of MJSynth \cite{jaderberg2014synthetic} and SynthText \cite{gupta2016synthetic} datasets. After filtering out words containing non-alphanumeric characters, these datasets contain 8.9M and 5.5M word-level images for training, respectively.

The related methods include \cite{bluche2017scan,kang2021candidate,kang2018convolve,michael2019evaluating,sueiras2018offline}, where \cite{bluche2017scan} is the first attempt towards using an attention-based model for HTR. The limitation of this work is that it requires pre-training of the features extracted from the encoding stage using the Connectionist Temporal Classification (CTC) loss to be relevant. The method proposed in \cite{sueiras2018offline} is based on a sliding window approach. Limitations of this work include an increase in overhead due to sliding window size initialization and parameter tuning. The model proposed in \cite{kang2018convolve} consists of an encoder, an attention mechanism, and a decoder. The task of the encoder is to extract deep visual features and it is implemented as a combination of two neural networks: a Convolutional Neural Network (CNN) and a bidirectional Gated Recurrent Unit (BGRU). The task of the decoder is to predict one character at a time. Each time a character is being predicted an attention mechanism automatically defines a context as a combination of input features that are the most relevant for the current predicted character and improves the predictive performance of the decoder. Similar to the proposed method, \cite{kang2018convolve} does not require pre-processing, a pre-defined lexicon, or a language model. The method proposed in \cite{michael2019evaluating} combines CNN and a deep bidirectional LSTM, and also investigates various attention mechanisms and positional encodings to address the problem of input-output text sequence alignment. For decoding the actual character sequence, it uses a separate RNN. Recently, \cite{kang2021candidate} introduced Candidate Fusion as a novel approach toward integrating a language model into a seq2seq architecture. The proposed method performs well in comparison with the related methods using the IAM dataset under the same experimental settings.

%to encode both the visual information, and the temporal context between characters in the input image. 

\section{Attention-based encoder-decoder network}
\label{attentionBasedEncoderDecoder}
%\label{sec:method}
%The overall pipeline of the proposed HTR method is presented in Fig. \ref{fig:archFull}, and will be studied in detail in this section. 

\begin{figure}[!t]
\centering
\includegraphics[height=3.1in]{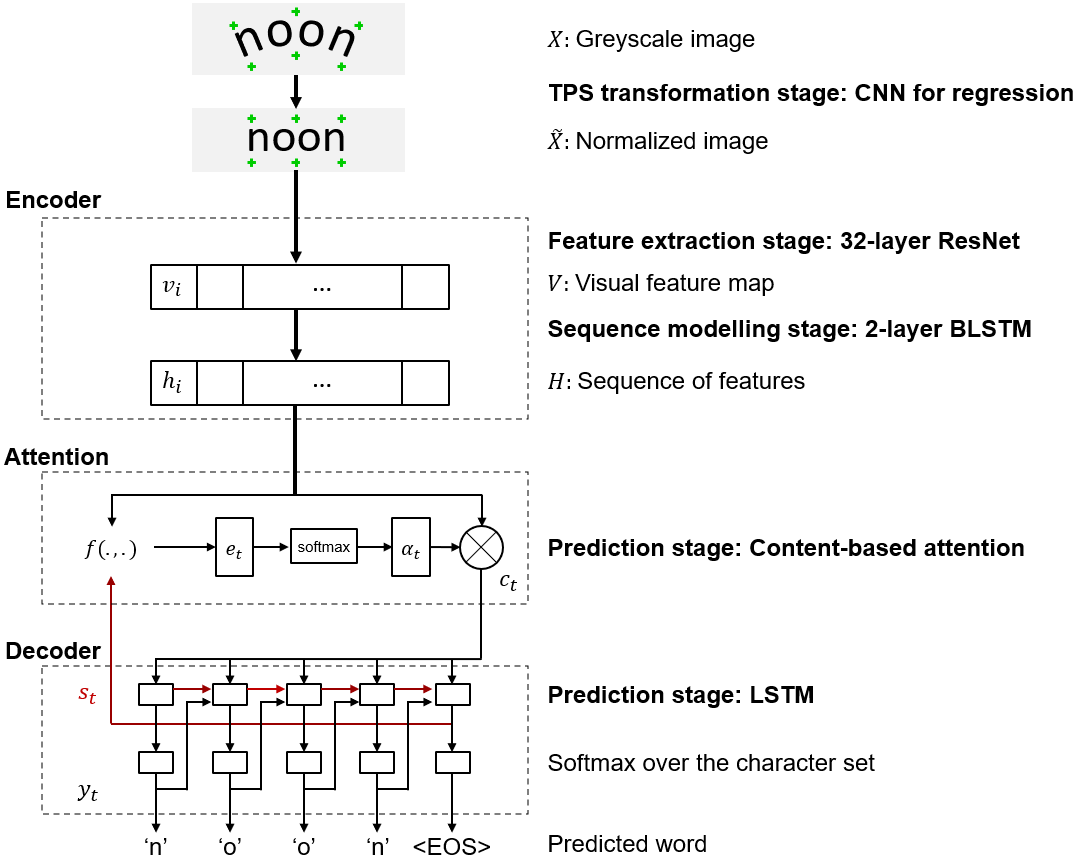}
\caption{Attention-based encoder-decoder architecture for HTR.}
\label{fig:archFull}
\end{figure}

%\subsection{Attention-based encoder-decoder} %\label{attentionBasedEncoderDecoder}
The overall pipeline of the proposed HTR method is presented in Fig. \ref{fig:archFull}. The model architecture consists of four stages: transformation, feature extraction, sequence modeling, and prediction, discussed as follows.

%Input is a 100$\times$30 greyscale image. First, it undergoes a spatial transformation stage, which aims at normalizing the text into a regular shape. Then feature extraction and sequence modeling stages together form an encoder and produce features used for predictions. The prediction stage consists of a decoder and a content-based attention mechanism. This final stage outputs a variable-length predicted word subject to a maximum length and uses characters from a predefined character set. Both maximum length and a character set are hyper-parameters, and two different character label spaces were used. Four stages can be combined in a single system and trained in an end-to-end manner, further discussed as follows.

\textbf{Transformation stage.} Since the handwritten words appear in irregular shapes (e.g. tilted, skewed, curved), the input word images are normalized using the thin-plate spline (TPS) transformation \cite{bookstein1989principal}. The localization network of TPS takes an input image and learns the coordinates of fiducial points that are used to capture the shape of the text. Coordinates of fiducial points are regressed by a CNN and the number of points is a hyper-parameter. TPS also consists of a grid generator that maps the fiducial points on the input image $\mathbf{X}$ to the normalized image $\mathbf{\Tilde{X}}$, and then a sampler interpolates pixels from the input to the normalized image.

%\begin{figure}[!t]
%\includegraphics[width=3.4in]{tps-samples}
%\caption{Sample images from the Imgur5K dataset before and after the TPS transformation. Original images were first converted into grayscale and rescaled to 100$\times$30.} \label{fig:tps}
%\end{figure}

\textbf{Feature extraction stage.} A 32-layer residual neural network (ResNet) \cite{he2016deep} is used to encode a normalized 100$\times$32 greyscale input image into a 2D visual feature map $\mathbf{V}=\{\mathbf{v}_i\}, i=1..I$, where $I$ is the number of columns in the feature map. The output visual feature map has 512 channels $\times$ 26 columns. Each column of the feature map corresponds to a receptive field on the transformed input image. Columns in the feature map are in the same order as receptive fields on the image from left to right \cite{shi2016end}.

\textbf{Sequence modeling stage.} The features $\mathbf{V}$ from the feature extraction stage are reshaped into sequences of features $\mathbf{H}$, where each column in a feature map $\mathbf{v}_i \in \mathbf{V}$ is used as a sequence frame \cite{shi2016end}. A bidirectional LSTM (BLSTM) is used for sequence modeling which allows contextual information within a sequence (from both sides) to be captured, and renders the recognition of character sequences simpler and more stable as compared to recognizing each character independently. For example, a contrast between the character heights in ``il'' can help recognize them correctly, as opposed to recognizing each character independently \cite{shi2016end}. Finally, two BLSTMs are stacked to learn a higher level of abstractions. Dimensions of output feature sequence are 256$\times$26.

\textbf{Prediction stage.} An attention-based decoder is used to improve character sequence predictions. The decoder is a unidirectional LSTM and attention is content-based. The prediction loop has $T$ time steps, which is the maximum length of a predicted word. At each time step $t=1..T$ the decoder calculates a probability distribution over the character set $$\mathbf{y}_t = \text{softmax}(\mathbf{W}_0 \mathbf{s}_t + \mathbf{b}_0)$$
where $\mathbf{W}_0$ and $\mathbf{b}_0$ are trainable parameters and $\mathbf{s}_t$ is a hidden state of decoder LSTM at time $t$. A character with the highest probability is used as a prediction. The decoder can generate sequences of variable lengths, but the maximum length is fixed as a hyper-parameter. We used the maximum length of 26 characters, including an end-of-sequence (EOS) token that signals the decoder to stop making predictions after EOS is emitted. The character set is also fixed. A case-insensitive model used further in experiments uses an alphanumeric character set of length 37, which consists of 26 lower-case Latin letters, 10 digits, and an EOS token. A character set of a case-sensitive model also includes 26 upper-case Latin letters and 32 special characters (\textasciitilde \textasciicircum \textbackslash $!"\#\$\%\&'()*+,-./:;<=>?@[]_`\{|\}$), and has a total length of 95 characters.

A hidden state of the decoder LSTM $\mathbf{s}_t$ is conditioned on the previous prediction $\mathbf{y}_{t-1}$, a context vector $\mathbf{c}_t$ and a previous hidden state $$\mathbf{s}_t = LSTM(\mathbf{y}_{t-1}, \mathbf{c}_t, \mathbf{s}_{t-1}).$$
A context $\mathbf{c}_t$ is calculated as a weighted sum of encoded feature vectors $\mathbf{H}=\mathbf{h}_1,\dots,\mathbf{h}_I$ from the sequence modeling stage as $$\mathbf{c}_t = \sum_{i=1}^{I} \alpha_{ti} \mathbf{h}_i,$$
where $\alpha_{ti}$ are normalized attention scores, also called attention weights or alignment factors, and are calculated as 
$$\alpha_{ti} = \frac{\text{exp}(e_{ti})}{\sum_{k=1}^{I} \text{exp}(e_{tk})},$$
where $e_{ti}$ are attention scores calculated using Bahdanau alignment function \cite{bahdanau2014neural}
$$e_{ti} = f(\mathbf{s}_{t-1}, \mathbf{h}_i) = \mathbf{v}^\top \text{tanh}(\mathbf{W} \mathbf{s}_{t-1} + \mathbf{V} \mathbf{h}_i + \mathbf{b}),$$
where $\mathbf{v}$, $\mathbf{W}$, $\mathbf{V}$ and $\mathbf{b}$ are trainable parameters. \\

%\subsection{Transfer learning} 
%\label{TheoryFineTuning}

\textbf{Transfer learning}. Transfer learning is used in this work to resolve the problem of insufficient training data. In general, it relaxes the assumption that the training and test datasets must be identically distributed \cite{tan-Deep-TL-survey} and tries to transfer the learned features from the source domain to the target domain. The source domain here is STR, and the target domain is HTR. This paper uses the strategy of fine-tuning \cite{tan-Deep-TL-survey} by initializing the weights from a pre-trained model and continuing the learning process by updating all 49.6M parameters in it. The architecture of the fine-tuned model is the same as that of the pre-trained model. This, together with a significant overlap between the source and the target domains (synthetic scene text and handwritten text), allows the use of training datasets that are relatively small in size (up to 213K), compared to the training dataset used to obtain the pre-trained model (14.4M).

The learning rate is an important parameter to consider in fine-tuning, and its impact on the results is examined. A high learning rate can make the model perform poorly by completely changing the parameters of a pre-trained model. And a low learning rate can adjust a pre-trained model's parameters to a new dataset without making many adjustments to the original weights. A batch size of 32 has been selected based on the size of the training set and the memory requirements. Early stopping is used as a regularization method to avoid over-fitting due to its effectiveness and simplicity. 

%\subsection{Regularization} 
%\label{sectEarly}
%\textbf{Regularization}. This work uses early stopping as a regularization method to avoid over-fitting due to its effectiveness and simplicity. In general, early stopping stores a copy of the model's parameters every time the validation set error decreases. When the training process stops, the stored parameters are returned instead of the latest ones. The training process stops when the validation set error does not improve for a predefined number of epochs, often called as patience.

\section{Experimental results}
\label{sec:exp}
%This section details the fine-tuning experiments with the two pre-trained models from \cite{baek2019wrong}, i.e. the case-sensitive and the case-insensitive benchmark models. The pre-trained models and source code for the STR framework \cite{baek2019wrong} is available at GitHub \footnote{\href{https://github.com/clovaai/deep-text-recognition-benchmark}}, including the fine-tuning procedure. In order to early-stop the training process to prevent over-fitting, early stopping has also been added.

%The GitHub repository\footnote{\href{https://github.com/clovaai/deep-text-recognition-benchmark}{https://github.com/clovaai/deep-text-recognition-benchmark}} of \cite{baek2019wrong} contains links to pre-trained models and the source code for the four-stage STR framework, including the fine-tuning procedure. This section details the fine-tuning experiments with the two of their pre-trained models, i.e. the case-sensitive and the case-insensitive benchmark models. In order to early-stop the training process to prevent over-fitting, early stopping was added by integrating a modified version of the source code available at Bjarte Sunde's GitHub repository\footnote{\href{https://github.com/Bjarten/early-stopping-pytorch}{https://github.com/Bjarten/early-stopping-pytorch}}.

%\textbf{Datasets.} 
\subsection{Datasets}

Two word-level multi-writer datasets, IAM \cite{marti2002iam} and Imgur5K \cite{krishnan2021textstylebrush}, are used. Raw image files were converted into a Lightning Memory-Mapped Database\footnote{\href{https://lmdb.readthedocs.io/}{https://lmdb.readthedocs.io/}} (LMDB) before the experiments. The main benefit of LMDB is an extremely fast speed of read transactions, which is important for the training process.

The \textbf{IAM} handwriting database \cite{marti2002iam} contains 1,539 handwritten forms, written by 657 authors, and the total number of words is 115,320, with a lexicon of approximately 13,500 words. This work uses the most widespread RWTH Aachen partition into writer-exclusive training, validation, and test sets \cite{kang2018convolve} to ensure a fair comparison with the related works. 

\begin{figure}[!t]
\centering
\includegraphics[width=0.6\textwidth]{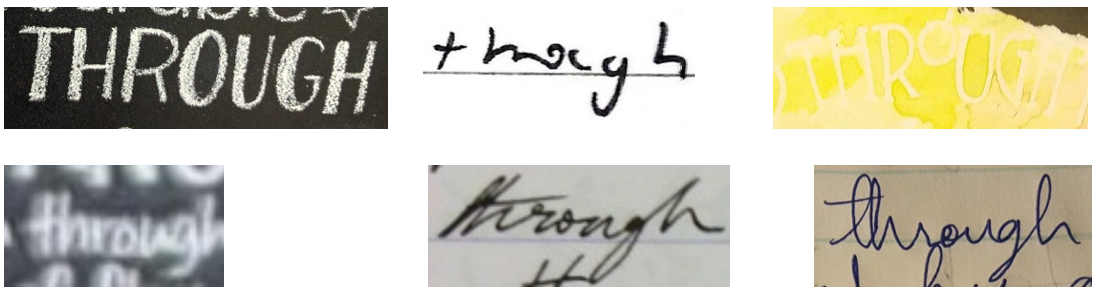}
\caption{\label{fig:imgur5k-examples}Imgur5K dataset samples. Top row: training set. Bottom row: test set.}
\end{figure}

%Partitions available on their GitHub repository\footnote{\href{https://github.com/omni-us/research-seq2seq-HTR}} are used to ensure consistency in result comparison. Only those words are kept, where segmentation of corresponding lines was marked as ``OK".
%\\

%\begin{figure}
%\centering
%\includegraphics[width=0.7\textwidth]{iam top train bottom test.png}
%\caption{\label{fig:iam-examples}Samples of cropped images containing a word ``through" from the IAM dataset. Top row: training set. Bottom row: test set.}
%\end{figure}

The \textbf{Imgur5K} dataset \cite{krishnan2021textstylebrush} is a new multi-writer dataset that contains 8,094 images, with the handwriting of approximately 5,305 authors published on an online image-sharing and image-hosting service \href{www.imgur.com}{imgur.com}. The total number of words is 135,375 with a lexicon of approximately 27,000 words. In comparison with the IAM dataset, the Imgur5K dataset contains 8 times more authors, 1.2 times more words, and 2 times larger lexicon. It is also a more challenging dataset in terms of background clutter and variability in pen types and styles. This work uses the same partitioning into document-exclusive training, validation, and test sets, as was provided from the dataset source. Fig. \ref{fig:imgur5k-examples} shows examples of a word \textit{through} in both training and test sets.

\begin{table}[htbp]
  \centering
  \caption{\label{tab:dataset-size}Size of training, validation, and test partitions used in experiments.}
    \begin{tabular}{llrrr}
    \toprule
    Dataset & Character set & \multicolumn{1}{c}{Training} & \multicolumn{1}{c}{Validation} & \multicolumn{1}{c}{Testing} \\
    \midrule
    \multirow{3}[2]{*}{IAM} & All characters & 47,981 & 7,554 & 20,305 \\
      & Case-sensitive (incl. special characters) & 47,963 & 7,552 & 20,300 \\
      & Case-insensitive & 41,228 & 6,225 & 17,381 \\
    \midrule
    \multirow{3}[2]{*}{Imgur5K} & All characters & 182,528 & 22,470 & 23,663 \\
      & Case-sensitive (incl. special characters) & 164,857 & 19,932 & 21,509 \\
      & Case-insensitive & 138,577 & 16,823 & 18,110 \\
    \midrule
    \multirow{3}[2]{*}{Total} & All characters & 230,509 & 30,024 & 43,968 \\
      & Case-sensitive (incl. special characters) & 212,820 & 27,484 & 41,809 \\
      & Case-insensitive & 179,805 & 23,048 & 35,491 \\
    \bottomrule
    \end{tabular}%
\end{table}%

The actual size of partitions differs from full partitions if an annotation is missing, the length of the annotation exceeds 25 characters, or if it contains characters that are not included in the model's character set. Furthermore, the IAM partitions contain only those words where the segmentation of corresponding lines was marked as ``OK''. The size of unfiltered and filtered partitions for both case-sensitive and case-insensitive models is shown in Table \ref{tab:dataset-size}.

%\textbf{Hyper-parameters.} 
\subsection{Hyper-parameters}

AdaDelta optimizer is used with the hyper-parameters matching those in \cite{baek2019wrong}: learning rate $\eta$=1; decay constant $\rho$=0.95; numerical stability term $\epsilon$=$1\mathrm{e}{-8}$; gradient clipping is 5; maximum number of iterations 300,000. Early stopping patience is set to 5 epochs, and training batch size is 32.

%htbp
\begin{table}[!t]
  \centering
  \caption{Validation set errors for the proposed method.}
    \begin{tabular}{cp{7.5em}p{4em}p{4em}p{4em}p{4em}p{4em}p{4em}}
    \toprule
    \multicolumn{1}{c}{\multirow{2}[4]{*}{{\parbox{0.9cm}{\centering Char. set}}}} & \multirow{2}[4]{*}{Fine-tuned on} & \multicolumn{3}{c}{Val-CER} & \multicolumn{3}{c}{Val-WER} \\
\cmidrule{3-8}      & \multicolumn{1}{c}{} & \multicolumn{1}{c}{Imgur5K} & \multicolumn{1}{c}{IAM} & Both & \multicolumn{1}{c}{Imgur5K} & \multicolumn{1}{c}{IAM} & Both \\
    \midrule
    \multirow{6}[2]{*}{{\rotatebox[origin=c]{90}{Case-insensitive}}} & Benchmark & 22.66 & 25.65 & 23.47 & 37.79 & 47.05 & 40.29 \\
      & IAM & 27.58 & 3.65 & 21.12 & 40.83 & 11.79 & 32.98 \\
      & IAM$\rightarrow$Imgur5K & \textbf{6.28} & 7.73 & 6.67 & \textbf{12.95} & 20.06 & 14.87 \\
      & Imgur5K & 6.70 & 8.48 & 7.18 & 13.48 & 22.43 & 15.90 \\
      & Imgur5K$\rightarrow$IAM & 13.69 & \textbf{2.73} & 10.73 & 23.43 & \textbf{8.76} & 19.47 \\
      & Imgur5K+IAM & 7.17 & 3.24 & \textbf{6.11} & 13.79 & 9.82 & \textbf{12.72} \\
      \addlinespace[1ex]
    \midrule
    \multirow{6}[2]{*}{{\rotatebox[origin=c]{90}{Case-sensitive}}} & Benchmark & 29.77 & 42.81 & 33.35 & 49.73 & 59.07 & 52.30 \\
      & IAM & 34.80 & 5.16 & 26.65 & 53.59 & 12.45 & 42.28 \\
      & IAM$\rightarrow$Imgur5K & \textbf{9.32} & 25.07 & 13.65 & \textbf{19.61} & 34.88 & 23.81 \\
      & Imgur5K & 9.18 & 23.31 & 13.06 & 19.74 & 36.32 & 24.29 \\
      & Imgur5K$\rightarrow$IAM & 22.43 & \textbf{4.88} & 17.61 & 38.48 & \textbf{11.80} & 31.15 \\
      & Imgur5K+IAM & 9.48 & 4.97 & \textbf{8.24} & 19.69 & 12.38 & \textbf{17.68} \\
      \addlinespace[1ex]
    \bottomrule
    \end{tabular}%
  \label{tab:valErrors}%
\end{table}%

%\subsection{Results}
%\label{sec:res}
%\textbf{Results.} 
\subsection{Results}
Two standard metrics: character error rate (CER) and word error rate (WER) are used to evaluate the experimental results. CER is defined as a normalized Levenshtein distance: $$CER = \frac{1}{N} \sum_{i=1}^{N} D(s_i, \hat{s}_i) / l_i,$$
where $N$ is the number of annotated images in the set, $D(.)$ is the Levenshtein distance, $s_i$ and $\hat{s}_i$ denote the ground truth text string and the corresponding predicted string, and $l_i$ is the length of the ground truth string. 
Given the predictions are evaluated at word level, the WER corresponds to the misclassification error: $$WER = \frac{1}{N} \sum_{i=1}^{N} \mathbbm{1}\{s_i \ne \hat{s}_i\}.$$

Table \ref{tab:valErrors} presents the validation set CER and WER. Models with a case-sensitive character set have 36 alphanumeric characters and a special EOS token. Models with a case-insensitive character set additionally have 26 upper-case letters and 32 special characters. In all experiments, training and validation sets are i.i.d. (independent and identically distributed). Sequential transfer learning using one and then the other dataset is denoted with a right arrow between the dataset names. Usage of two datasets as a union is denoted with a plus sign.

When transfer learning is performed sequentially, the lowest error on either Imgur5K or IAM is obtained when the corresponding dataset is used last in the fine-tuning sequence. The lowest error on the union of two datasets is obtained when both datasets are used for transfer learning at the same time. These models also have a competitive performance on stand-alone datasets. Furthermore, the results show that the model's domain of application determines the fine-tuning approach. If the unseen data is close to the IAM distribution then a fine-tuning sequence Imgur5K$\rightarrow$IAM is expected to give the lowest error. If the unseen data is close to the Imgur5K distribution then two datasets should be used in a reversed sequence. And a more general-purpose handwritten text recognition model, as represented by a union of the two datasets, is more likely to be obtained by fine-tuning on the union of the datasets. 

%\subsubsection{Hyper-parameter tuning}
%As an example, \textbf{hyper-parameter tuning} is performed for the case-insensitive benchmark model, fine-tuned on the Imgur5K$\rightarrow$IAM sequence. Two hyper-parameters are tuned: AdaDelta optizer's learning rate $\eta$ and decay constant $\rho$. AdaDelta's decay constant is similar to that used in the momentum method \cite{qian1999momentum}. 

\begin{table}[!t]
  \centering
  \caption{\label{tab:adadeltaHyper}Validation WER on varying the learning rate $\eta$ and decay constant $\rho$ for the case-insensitive benchmark model, fine-tuned on the Imgur5K$\rightarrow$IAM sequence. The lowest error is shown in \textbf{bold}. Cells with the lowest row-wise error have a grey fill.}
    \begin{tabular}{p{8em}>{\centering\arraybackslash}p{4em}>{\centering\arraybackslash}p{4em}>{\centering\arraybackslash}p{4em}}
    \toprule
    \multirow{2}[4]{*}{Learning rate $\eta$} & \multicolumn{3}{c}{Decay constant $\rho$} \\
\cmidrule{2-4}      & 0.9 & 0.95 & 0.99 \\
    \midrule
    1.5 & \cellcolor[rgb]{ .851,  .851,  .851}9.69 & 11.07 & 11.47 \\
    1 & 9.59 & \cellcolor[rgb]{ .851,  .851,  .851}\textbf{8.76} & 9.98 \\
    0.1 & 9.27 & \cellcolor[rgb]{ .851,  .851,  .851}8.96 & 8.96 \\
    0.01 & 9.30 & 9.49 & \cellcolor[rgb]{ .851,  .851,  .851}9.03 \\
    0.001 & 11.33 & 10.86 & \cellcolor[rgb]{ .851,  .851,  .851}10.67 \\
    \bottomrule
    \end{tabular}%
\end{table}%

%\textbf{Hyper-parameter tuning}. 
\subsection{Hyper-parameter tuning}

AdaDelta optimizer's learning rate $\eta$ and decay constant $\rho$ are tuned for the case-insensitive benchmark model fine-tuned on the Imgur5K$\rightarrow$IAM sequence. Table \ref{tab:adadeltaHyper} summarizes the WER on the IAM validation set. The lowest error corresponds to the default $\eta=1$ and $\rho=0.95$. As the learning rate increases from 0.001 to 1.5, lower error values correspond to lower decay constants. This indicates that large steps in a gradient descent work better when less weight is given to the last gradient. This observation is consistent with \cite{smith2018disciplined} that suggests a large momentum (i.e., 0.9-0.99) with a constant learning rate can act as a pseudo increasing learning rate, and speed up the training. Whereas oversize momentum values can cause poor training results as the parameters of the pre-trained model get too distorted.

\begin{figure}[!t]
\centering
\includegraphics[width=0.7\textwidth]{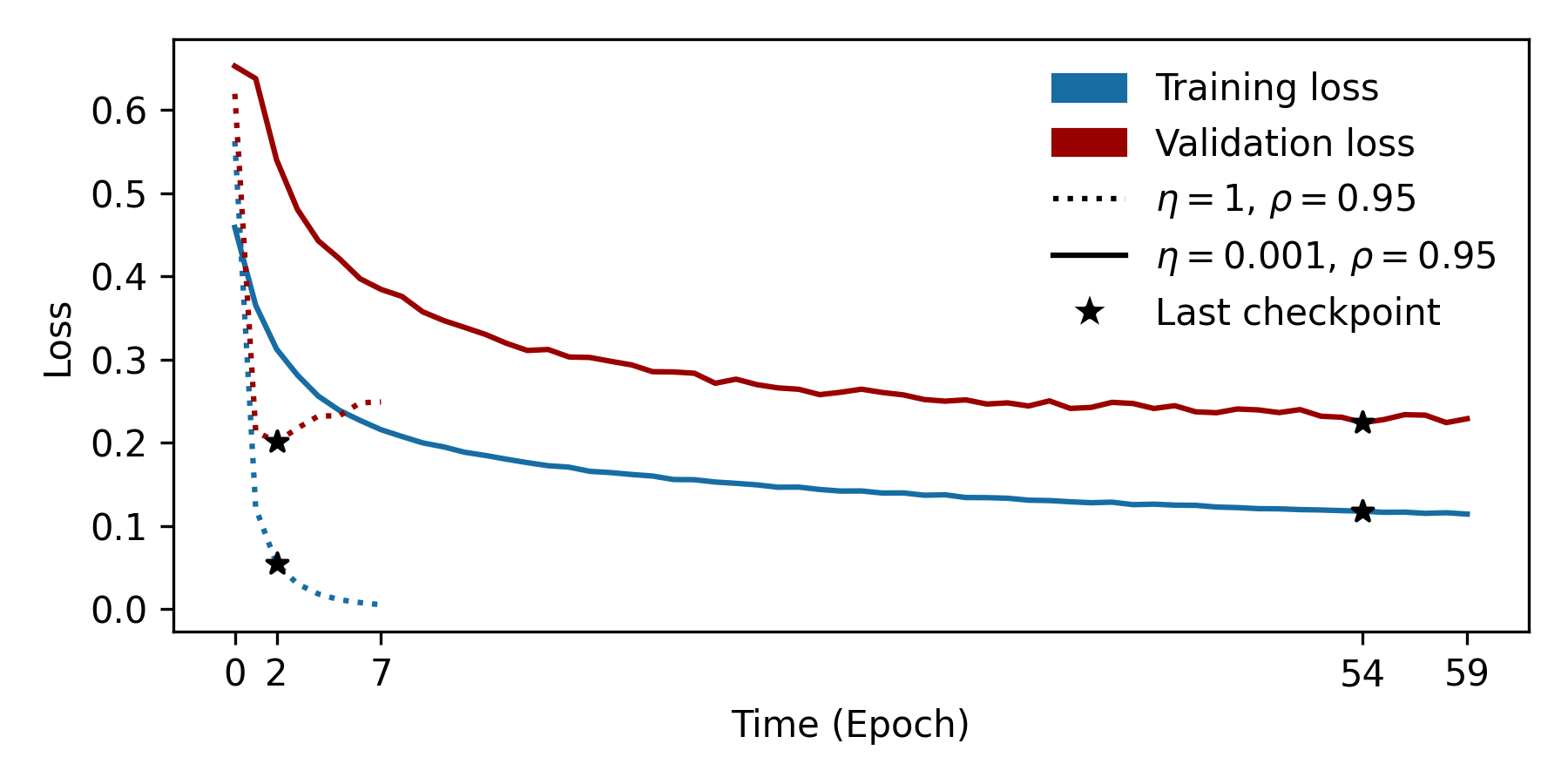}
\caption{\label{fig:compare2134}Comparison of training and validation loss for the case-insensitive benchmark model fine-tuned on the Imgur5K$\rightarrow$IAM sequence using two sets of hyper-parameters.}
\end{figure}

%A too low patience value of the early stopping mechanism leads to an increased chance of under-fitting when combined with a low learning rate. Solid lines represent training (blue color) and validation (red color) losses with the learning rate $\eta=0.001$ and dotted lines represent $\eta=1$.

The IAM dataset is relatively small with only 41K training images, compared to 14.4M training images of the corresponding benchmark model, and it is therefore recommended to perform fine-tuning with low learning rates. However, a low learning rate $\eta=0.001$ led to relatively high errors for all values of the decay constant. Fig. \ref{fig:compare2134} highlights that training with $\eta=0.001$ was stopped when its training loss (solid blue line) was significantly above the training loss achieved with $\eta=1$ (dotted blue line). This indicates an under-fitting problem caused by the early stopping mechanism. By switching the early stopping mechanism off and training for the full 300,000 iterations, we can achieve an error reduction from 10.86\%$\rightarrow$9.4\% in validation WER. In future work, other regularization strategies will be investigated such as the norm penalty or dropout. 

%We conclude that an early stopping mechanism is not an optimal regularization strategy and other forms of regularization, such as the norm penalty or dropout, need to be tried.

% \begin{figure}[!t]
% \centering
% \includegraphics[width=3.4in]{ft-seed-compare-models-no_early-stopping}
% \caption{\label{fig:longTraining}Model 20 fine-tuned on the IAM dataset with $\eta=0.001$ and $\rho-0.95$ for X epochs without early stopping. Figure best viewed in color.}
% \end{figure}

% \textbf{Ablation study}. The ablation study was performed to determine the importance of the following two components in the proposed framework:

% \begin{enumerate}
%     \item Transfer learning, as opposed to random initialization of the model's parameters.
%     \item Attention mechanism, as opposed to using CTC in the prediction stage.
% \end{enumerate}

%\textbf{Ablation study}. 
\subsection{Ablation study}
\begin{figure}[!t]
\centering
\includegraphics[width=1\textwidth]{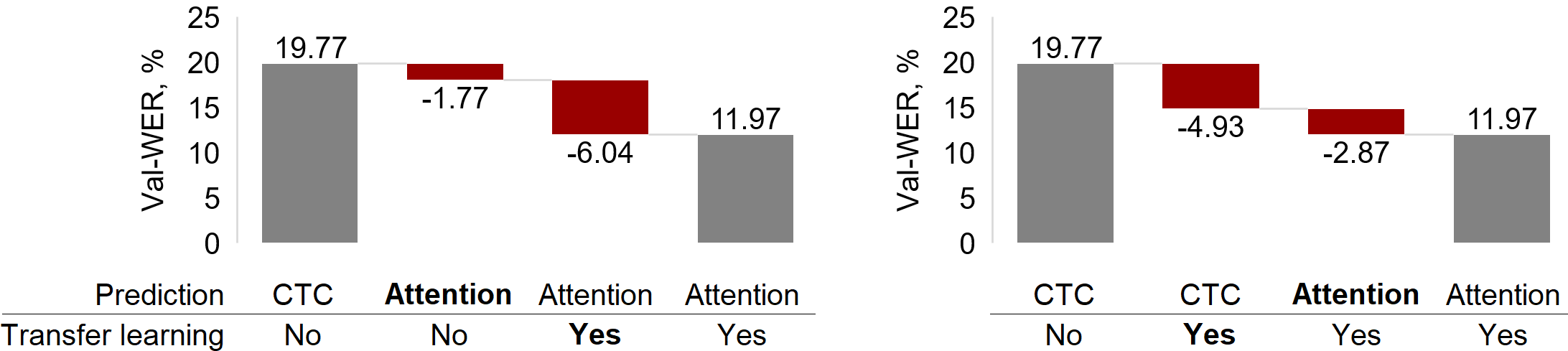}
\caption{\label{fig:ablation}The effect of transfer learning and attention mechanism on validation set WER for the IAM dataset when attention mechanism is introduced first (left), and when transfer learning is introduced first (right). Change is highlighted with \textbf{bold}.}
\end{figure}

An ablation study is conducted to demonstrate the effectiveness of transfer learning and attention mechanism for the given problem, and also to analyze the impact on the overall performance on the IAM dataset. This study is performed using a case-sensitive model trained and validated on the IAM dataset, and each model was trained five times using different random seeds. Table \ref{fig:ablation} presents the average validation set CER and WER. When transfer learning from the STR domain and the Imgur5K dataset is not used, the model's parameters are initialized randomly (indicated as "None" in the transfer learning column). When the attention mechanism is not used for prediction, it is replaced by the CTC.

%The importance of the proposed framework based on transfer learning and the attention mechanism is demonstrated with an ablation study using a case-sensitive model, trained and validated on the IAM dataset, as an example. Each model was trained five times using different random seeds. Average validation set CER and WER are presented in Table \ref{fig:ablation}. When transfer learning from the STR domain and the Imgur5K dataset is not used, the model's parameters are initialized randomly (indicated as "None" in the transfer learning column). When the attention mechanism is not used for prediction, it is replaced by the CTC. The model with random initialization of parameters and CTC for prediction is used as a baseline. The numbers in parentheses show a reduction in errors again this baseline. The lowest errors and corresponding components are shown in \textbf{bold}.

\begin{table}[!t]
  \centering
  \caption{\label{tab:ablation}Ablation study highlighting the importance of transfer learning and attention mechanism. The lowest errors and corresponding components are shown in \textbf{bold}.}
    \begin{tabular}{p{3cm}p{3cm}cc}
    \toprule
    Transfer learning & Prediction & Val-CER, \% & Val-WER, \% \\
    \midrule
    None  & CTC   & 7.12  & 19.77 \\
    None  & Attention & 6.79 & 18.01 \\
    STR, Imgur5K & CTC   & 5.32 & 14.84\\
    \textbf{STR, Imgur5K} & \textbf{Attention} & \textbf{4.84}  & \textbf{11.97}\\
    \bottomrule
    \end{tabular}%
\end{table}%

%\begin{table}[!t]
%  \centering
%  \caption{\label{tab:ablation}Ablation study summary showing the importance of transfer learning and the attention mechanism. Each model was trained five times and errors were averaged. Changes in errors are presented in parentheses and use the first line as a baseline.}
%    \begin{tabular}{cllcc}
%    \toprule
%    \#    & Transfer learning & Prediction & Val-CER, \% & Val-WER, \% \\
%    \midrule
%    1     & None  & CTC   & 7.12  & 19.77 \\
%    2     & None  & Attention & 6.79 (-0.33) & 18.01 (-1.77) \\
%    3     & STR, Imgur5K & CTC   & 5.32 (-1.80) & 14.84 (-4.93) \\
%    4     & \textbf{STR, Imgur5K} & \textbf{Attention} & \textbf{4.84} (-2.28) & \textbf{11.97} (-7.81) \\
%    \bottomrule
%    \end{tabular}%
%\end{table}%

The ablation study proves that transfer learning and attention mechanism help reduce both CER and WER. The lowest errors are achieved when both transfer learning and attention mechanism are used. The effect of each component depends on which of them comes first, as illustrated in Fig. \ref{fig:ablation}. Interestingly, in both cases, transfer learning is relatively more important among the two studied components in terms of error reduction.

%each of them is introduced separately in a sequence. The left figure shows the reduction in WER when the attention mechanism is introduced first. The right figure shows the reduction in WER when transfer learning is introduced first.

% Ablation study comparing case-sensitive models trained and tested on the AIM dataset. Both figures start with no transfer learning and CTC for the prediction stage. And both end with transfer learning and attention-based prediction. The left figure shows how the validation WER changes if the prediction is changed from CTC to the attention mechanism first and transfer learning is introduced in the following step. In the right figure these two changes are introduced in the reversed order.

% \subsubsection{Test error}

%\textbf{Test set errors}. 
\subsection{Test set errors}
Table \ref{tab:testErrors} highlights the \textbf{test set CER and WER} for our best performing models on each dataset. A comparison is performed with related works on content attention-based seq2seq models, under the same experimental settings using the IAM dataset. The test set WER values suggest that the proposed method outperformed all the state-of-the-art methods \cite{kang2021candidate}, \cite{kang2018convolve}, \cite{sueiras2018offline}. The test set WER for the proposed method is 15.40\%. Kang et al. \cite{kang2021candidate} achieved a WER comparable to our work (15.91\%) using a content-based attention mechanism. Kang et al. \cite{kang2021candidate} also used location-based attention, achieving a WER of 15.15\%, but since it explicitly includes the location information into the attention mechanism, it is not considered in this work. Furthermore, \cite{kang2021candidate} incorporated a language model, unlike the proposed method. Though an RNN decoder can learn the relations between different characters in a word, it is worth investigating the use of language models to improve the performance of the proposed method as future work.
%Since the predictions are evaluated at the word level, a low WER value indicates a low misclassification error achieved by the proposed method. 

The test set CER results on the other hand suggest that Johannes et al. \cite{michael2019evaluating} outperformed all the state-of-the-art methods, with a test set CER of 5.24\%. Similar to the proposed work, \cite{michael2019evaluating} did not use a language model. However, Kang et al. \cite{kang2021candidate} improved upon their previous work \cite{kang2018convolve} by integrating a language model, and achieved a low error rate (5.79\%). The proposed method achieved 6.50\% CER and performs comparably with \cite{michael2019evaluating} and \cite{kang2021candidate}, outperforming the rest of the methods. Based on our error analysis (discussed in detail in the next section), the CER can be further reduced by using data augmentation, language modeling, and a different regularization method.

We also performed tests on the case-insensitive model that achieved 4.3\% CER and 12.82\% WER on IAM. This work is also the first attempt at using the Imgur5K dataset for handwritten word recognition (to the best of the authors' knowledge), and therefore no related work exists to compare the performance on the Imgur5K dataset. However, it can be observed from Table \ref{tab:testErrors} that the proposed model has a relatively low test set CER and WER for Imgur5K, and can model a variety of handwritten words from this novel multi-writer dataset.

In summary, the proposed method has the lowest error rate when evaluated at the word level, and the third-lowest error rate when evaluate at the character level. The advantages of using the proposed approach in comparison with \cite{michael2019evaluating} and \cite{kang2021candidate} include addressing the problem of training data scarcity, performing transfer learning from STR to HTR, and using Imgur5K dataset with word images from 5000 different writers. This allows the proposed HTR system to also handle examples that are rare due to insufficient annotated data. Our models are publicly available for further fine-tuning or predictions on unseen data.

\begin{table}[!t]
  \centering
  \caption{\label{tab:testErrors}Results comparison with the state-of-the-art content attention-based seq2seq models for handwritten word recognition.}
    \begin{tabular}{lcccccc}
    \toprule
    \multicolumn{1}{c}{\multirow{2}[4]{*}{Character set}} & \multicolumn{3}{p{12em}}{\centering Test-CER} & \multicolumn{3}{p{12em}}{\centering Test-WER} \\
\cmidrule{2-7}      & \multicolumn{1}{c}{Imgur5K} & \multicolumn{1}{c}{IAM} & \multicolumn{1}{c}{Both} & \multicolumn{1}{c}{Imgur5K} & \multicolumn{1}{c}{IAM} & \multicolumn{1}{c}{Both} \\
    \midrule
    \textbf{Ours}, case-insensitive & \multicolumn{1}{c}{6.46$^a$} & 4.30$^b$ & \multicolumn{1}{c}{5.96$^c$} & \multicolumn{1}{c}{13.58$^a$} & 12.82$^b$ & \multicolumn{1}{c}{13.89$^c$} \\
    \midrule
    \textbf{Ours}, case-sensitive & \multicolumn{1}{c}{9.47$^a$} & 6.50$^b$ & \multicolumn{1}{c}{8.59$^c$} & \multicolumn{1}{c}{20.45$^a$} & \textbf{15.40}$^b$ & \multicolumn{1}{c}{18.97$^c$} \\
    %\midrule
    Kang \textit{et al.} \cite{kang2018convolve} & \multicolumn{1}{c}{-} & 6.88 & \multicolumn{1}{c}{-} & \multicolumn{1}{c}{-} & 17.45 & \multicolumn{1}{c}{-} \\
    Bluche \textit{et al.} \cite{bluche2017scan} & \multicolumn{1}{c}{-} & 12.60 & \multicolumn{1}{c}{-} & \multicolumn{1}{c}{-} & - & \multicolumn{1}{c}{-} \\
    Sueiras \textit{et al.} \cite{sueiras2018offline} & \multicolumn{1}{c}{-} & 8.80 & \multicolumn{1}{c}{-} & \multicolumn{1}{c}{-} & 23.80 & \multicolumn{1}{c}{-} \\
    Chowdhury \textit{et al.} \cite{chowdhury2018efficient} & \multicolumn{1}{c}{-} & 8.1 & \multicolumn{1}{c}{-} & \multicolumn{1}{c}{-} & - & \multicolumn{1}{c}{-} \\
    Johannes \textit{et al.} \cite{michael2019evaluating} & \multicolumn{1}{c}{-} & \textbf{5.24} & \multicolumn{1}{c}{-} & \multicolumn{1}{c}{-} & - & \multicolumn{1}{c}{-} \\
    Kang \textit{et al.} \cite{kang2021candidate}$^d$ & 
    \multicolumn{1}{c}{-} & 5.79 & \multicolumn{1}{c}{-} & \multicolumn{1}{c}{-} & 15.91 & \multicolumn{1}{c}{-} \\
    \bottomrule
    \multicolumn{7}{l}{%
  \begin{minipage}{12cm}%
    \footnotesize Fine-tuning approaches: $^a$ IAM$\rightarrow$Imgur5K; $^b$ Imgur5K$\rightarrow$IAM; $^c$ Imgur5K+IAM. \\$^d$Also provides the results using location-based attention mechanism, which is not applicable in the considered content-based attention experimental settings.
  \end{minipage}}
    \end{tabular}%
%\begin{tablenotes}
%   \item[*] $^a$ IAM$\rightarrow$Imgur5K; $^b$ Imgur5K$\rightarrow$IAM; $^c$ Imgur5K+IAM; $^d$ Re-trained on the training and validation sets; $^e$ Uses a language model.
%  \end{tablenotes}
\end{table}%

\section{Error analysis} \label{sectErAnal}
\label{sec:errorAnalysis}

This section performs error analysis for our case-insensitive model performing best on the union of Imgur5K and IAM datasets. 

\subsection{Character-level error analysis} \label{errorAnalConfMat}

% This section analyses precision and recall of the case-insensitive model on a character level. The model being analyzed is our best performing model fine-tuned on the union of the Imgur5K and IAM datasets. 

\begin{figure}[!t]
\centering
\includegraphics[width=1\textwidth]{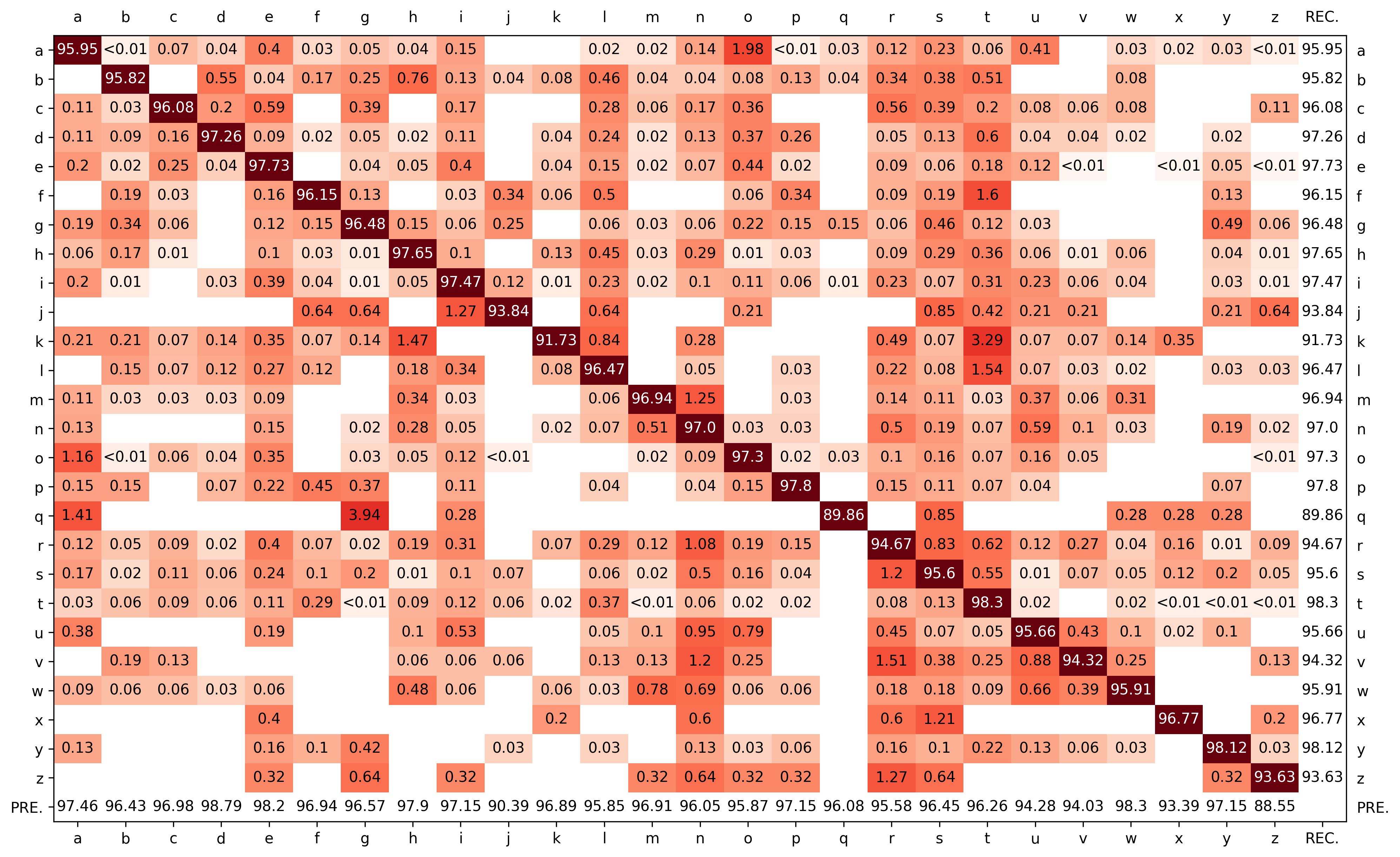}
\caption{\label{fig:confMatNorm}Character-level error analysis using a confusion matrix with a row-wise normalization. The main diagonal represents recall. Each row adds up to 1, possibly with a rounding error. Abbreviations used: PRE. stands for precision, REC. stands for recall.}
\end{figure}

\begin{figure}[!t]
\centering
\includegraphics[height=1.7in]{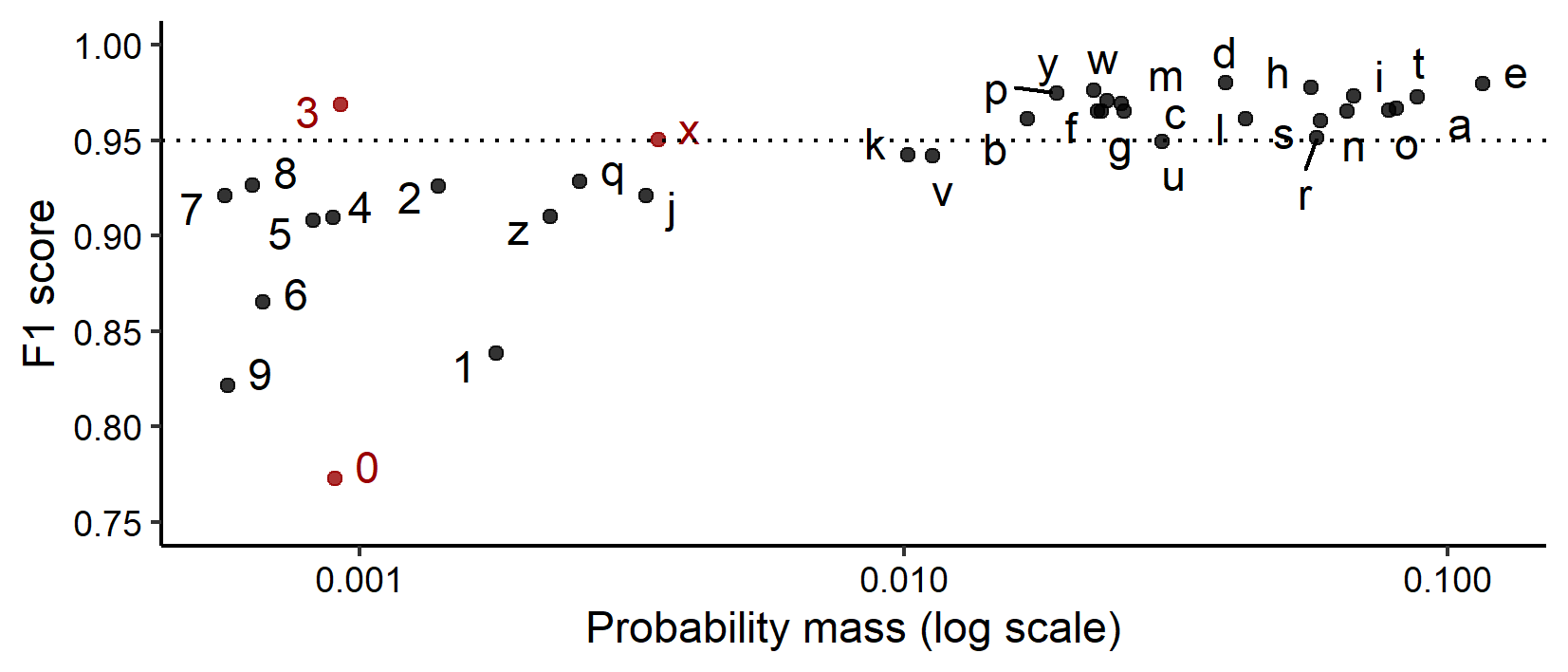}
\caption{\label{fig:f1}F1 scores vs probability mass of lower-case alphanumeric characters. Characters discussed in the text are depicted in red. Figure best viewed in color.}
\end{figure}

To evaluate the model's predictive performance at the character level, precision and recall are computed from a confusion matrix. It is based on 33,844 images from 35,491 images in the test set where the length of the predicted word matches the length of the ground truth word. This is necessary to have 1-to-1 correspondence between true and predicted characters. Fig. \ref{fig:confMatNorm} presents a normalized confusion matrix with the ground truth characters in rows and predicted characters in columns. Only letters are shown for better readability. Values in each row are divided by the total number of characters in that row. Therefore, the main diagonal contains recall values, and each row adds up to 1, possibly with a rounding error. Off-diagonal values represent errors. The recall is also shown on the right margin and precision is shown on the bottom margin. All values are in percent, rounded to two decimal digits. Additionally, precision and recall are combined into a single metric, F1 score, and plotted against the probability mass (normalized frequency) of each character on Fig. \ref{fig:f1}. These two figures present several interesting and valuable insights: 

\begin{itemize}
    \item The model's errors pass a sanity check as the confusion matrix highlights that the most typical errors are between characters having similar ways of writing. For example, \textit{a} is most often confused with \textit{o}, and \textit{q} with \textit{g}.
    \item Letters \textit{z}, \textit{q}, \textit{j} and \textit{x} and all digits are significantly under-represented in the distribution of characters.
    \item In general, a lower probability mass of a character leads to a lower F1 score. However, the uniqueness of a character's way of writing seems to be an important factor. For example, digits \textit{0} and \textit{3} have approximately the same probability mass but hugely different F1 scores (77\% and 97\%). This difference is attributed to the fact that digit \textit{3} is relatively unique and is, therefore, more difficult to confuse with other characters. The same conclusion applies to letter \textit{x}, which achieves a relatively high F1 score despite being among the least frequent characters. Intuitively, characters that are somewhat unique in the way they are written need fewer examples in the training set to achieve high F1 scores. This also suggests that F1 scores can be used to prioritize characters for inclusion in the data augmentation, as a possible next step to improve the model's performance.
\end{itemize}

\subsection{Bias-variance analysis} \label{sectBiasVariance}

To gauge the model's performance, bias-variance analysis is performed according to \cite{AndreNgErrorAnalysis}. It yields only approximate values of the test error components but is helpful and instructive from a practical point of view. Human-level error is used as a proxy for Bayes error, which is analogous to irreducible error and denotes the lowest error that can be achieved on the training set before the model's parameters start over-fitting. To estimate a human-level error, 22 human subjects from different backgrounds were asked to annotate the same set of randomly selected 100 misclassified images from the test set. The images that are misclassified by \textit{all} reviewers contribute to the human-level error estimate.

\begin{figure}[!t]
\centering
\includegraphics[width=0.8\textwidth]{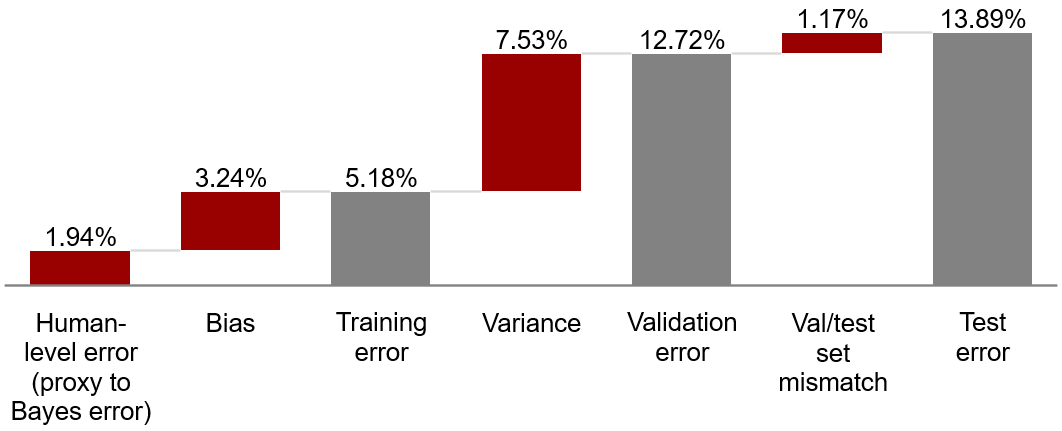}
\caption{\label{fig:waterfall}Bias-variance decomposition. Red bars are components of the test set WER on the union of Imgur5K and IAM. Grey bars are the model's errors obtained by summing red bars from left to right. Human-level error is used as a proxy to Bayes error.}
\end{figure}

In general, the bias of the model is the difference between Bayes error and the training set error. The variance is the difference between the training and validation set error. Additionally, the difference between the validation and test set error is denoted by the validation/test set mismatch error. Fig. \ref{fig:waterfall} summarizes the results of the bias-variance analysis. By definition, it is not feasible to reduce the Bayes error, which is 1.94\% according to our estimate. It can be observed in Fig. \ref{fig:waterfall} that the variance contributes the largest part of the error (7.53\%) among other components and should be the primary focus of measures aimed at decreasing the testing error of the model. Suitable strategies include using a larger training set, data augmentation, and a different regularization method (e.g. norm penalty, dropout) instead of early stopping. The second largest component is the bias (3.24\%) that can be reduced by, for example, training a bigger neural network, using a different optimizer (e.g. Adam), and running the training longer. The smallest component (1.17\%) is due to the validation/test set mismatch; this can be addressed by reviewing the partitioning strategy for the IAM dataset. However, this work follows the same splits as widely used in the HTR research for a fair comparison with other related work.

%To summarize, methods aimed at reducing the variance should receive the highest priority for the model fine-tuned on the union of Imgur5K and IAM datasets:

%%%%%%%%%%%%%%%%%%%%%%%%%%%%%%%%%%%%%%%
%%%%%%%%%%%%%%%% START 1 %%%%%%%%%%%%%%
%%%%%%%%%%%%%%%%%%%%%%%%%%%%%%%%%%%%%%%

%\begin{enumerate}
%    \item Replace early stopping-type of regularization with the norm penalty or dropout. 
%    \item Data augmentation. Data augmentation is another form of regularization \cite{Goodfellow-et-al-2016} \cite{AndreNgErrorAnalysis}. It can be implemented in at least two ways. First, by generating synthetic images. \cite{TextRecognitionDataGenerator} is an example of an open-source synthetic data generator for text recognition. Second, by modifying existing images from the training set. In \S\ref{ManualInspectionofImages} we arrive at an informed strategy for data augmentation.
%\end{enumerate}

%%%%%%%%%%%%%%%%%%%%%%%%%%%%%%%%%%%%%%%
%%%%%%%%%%%%%%%% END 1 %%%%%%%%%%%%%%%%
%%%%%%%%%%%%%%%%%%%%%%%%%%%%%%%%%%%%%%%

\subsection{Visual analysis of images} \label{ManualInspectionofImages}
\begin{figure}[!t]
\centering
\includegraphics[width=4.5in]{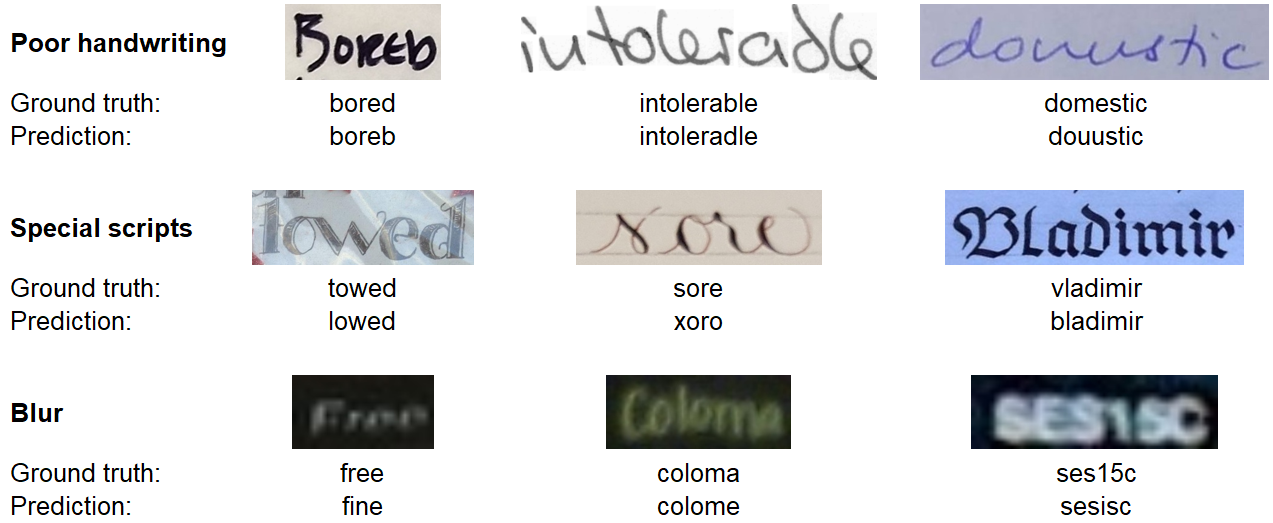}
\caption{\label{fig:VisualInspection}Sample images from selected categories. Each category represents a possible main cause for misclassification, as identified through visual analysis.}
\end{figure}

To provide further insights into the possible causes for the model's errors, a visual inspection is performed on the same set of 100 misclassified images. The main problems contributing towards errors include poor handwriting, the use of special scripts (e.g. Gothic, curlicue, etc.), errors in some ground truth images, and other forms of issues such as low image resolution, blurry and rotated text, challenging backgrounds, strike-through words, or multiple words in an image. Fig. \ref{fig:VisualInspection} contains sample images representing some of the listed problems. The analysis suggests that the performance of the model can be further improved using data augmentation (with special scripts, blur, rotations, variety of backgrounds, etc.), and also fine-tuning on other multi-writer datasets. A language model is also foreseen to reduce the error rate on images with poor handwriting.

%Result of a visual analysis can further clarify the tactics for improving the model's performance:

%%%%%%%%%%%%%%%%%%%%%%%%%%%%%%%%%%%%%%%
%%%%%%%%%%%%%%%% START 2 %%%%%%%%%%%%%%
%%%%%%%%%%%%%%%%%%%%%%%%%%%%%%%%%%%%%%%

%\begin{enumerate}
%    \item \textbf{Data augmentation}. Table \ref{tab:VisualInspection} allows for an informed design of data augmentation by focusing on the most common problems, such as special scripts and blur. Text rotation and a variety of background may receive less attention.
%    \item \textbf{Other annotated multi-writer datasets}. Fine-tuning on other datasets is expected to improve the model's ability to perform well on a wider variety of handwriting styles.
%    \item \textbf{Language model} is expected to reduce an error rate on images from the bad/wrong handwriting category.
%\end{enumerate}

\section{Conclusion}
\label{sec:conclusion}
This work presented an end-to-end-HTR system based on attention encoder-decoder networks, which leverages pre-trained models trained on a large set of synthetic scene text images for handwritten word recognition. The problem of training data scarcity is addressed by performing transfer learning from STR to HTR, and using a new multi-writer dataset (Imgur5K) that contains word examples from 5000 different writers. This allows the proposed HTR system to also cope with rare examples due to insufficient annotated data. The experimental results on multi-writer datasets (IAM and Imgur5K) demonstrate the effectiveness of the proposed method. Under the given experimental settings, the proposed method outperformed the state-of-the art methods by achieving the lowest error rate when evaluated at the word level on the IAM dataset (test set WER 15.40\%). At character level, the proposed method performed comparable with the state-of-the-art methods and achieved 6.50\% test set CER. However, the character level error can be further reduced by using data augmentation, language modeling, and a different regularization method, which will be investigated as future work. Our source code and pre-trained models are publicly available for further fine-tuning or predictions on unseen data at GitHub\footnote{\href{https://github.com/dmitrijsk/AttentionHTR}{https://github.com/dmitrijsk/AttentionHTR}}.

%This work presented an end-to-end-HTR system based on attention encoder-decoder networks, which leverages pre-trained models trained on synthetic scene text images for handwritten word recognition, and uses transfer learning from STR domain to HTR to overcome the issue of training data scarcity. The experimental results on multi-writer datasets (IAM and Imgur5K) demonstrate the effectiveness of the proposed method. The proposed model coupled with the transfer learning pipeline significantly improved the word recognition accuracy. Under the given experimental settings, the proposed method performed well in comparison with the related state-of-the-art methods on the IAM dataset, with good accuracy even on the challenging word samples from the Imgur5K dataset. Our models are publicly available for further fine-tuning or predictions on unseen data. As future work, more datasets will be added to the pipeline, and regularization strategy and language modeling will be investigated. 

%Our best performing model is a union of IAM and Imgur5K datasets that can be used for performing HTR in general. The source code is available at GitHub\footnote{\href{https://github.com/dmitrijsk/HTR-Attention-Encoder-Decoder}{https://github.com/dmitrijsk/HTR-Attention-Encoder-Decoder}}. As future work, more datasets will be added in the pipeline, and regularization strategy and language modeling will be investigated. 

\section*{Acknowledgment}
This work has been partially supported by the Riksbankens Jubileumsfond (Reference number IN20-0040, Labour's Memory project). The authors would like to thank the Centre for Digital Humanities Uppsala, SSBA, Anders Hast and Örjan Simonsson for their kind support and encouragement provided during the project. The computations were performed on resources provided by SNIC through Alvis @ C3SE under project SNIC 2021/7-47. 

\bibliographystyle{splncs04}
\bibliography{samplepaper}

%\begin{figure}
%\includegraphics[width=\textwidth]{fig1.eps}
%\caption{A figure caption is always placed below the illustration.
%Please note that short captions are centered, while long ones are
%justified by the macro package automatically.} \label{fig1}
%\end{figure}

%\begin{table}
%\caption{Table captions should be placed above the
%tables.}\label{tab1}
%\begin{tabular}{|l|l|l|}
%\hline
%Heading level &  Example & Font size and style\\
%\hline
%Title (centered) &  {\Large\bfseries Lecture Notes} & 14 point, bold\\
%1st-level heading &  {\large\bfseries 1 Introduction} & 12 point, bold\\
%2nd-level heading & {\bfseries 2.1 Printing Area} & 10 point, bold\\
%3rd-level heading & {\bfseries Run-in Heading in Bold.} Text follows & 10 point, bold\\
%4th-level heading & {\itshape Lowest Level Heading.} Text follows & 10 point, italic\\
%\hline
%\end{tabular}
%\end{table}

\end{document}